\documentclass[conference]{IEEEtran}
\usepackage[utf8]{inputenc}
\usepackage{inconsolata}
\usepackage{cite}
\usepackage{amsmath,amssymb,amsfonts}
\usepackage{algorithmic}
\usepackage{diagbox}
\usepackage{multirow}
\usepackage{graphicx}
\usepackage{textcomp}
\usepackage{listings}
\usepackage{xcolor}
\usepackage{inconsolata}
\usepackage{hyperref}

\hypersetup{
    colorlinks=true,
    linkcolor=blue,
    citecolor=blue,
    filecolor=magenta,
    urlcolor=cyan
}

\def\BibTeX{{\rm B\kern-.05em{\sc i\kern-.025em b}\kern-.08em
    T\kern-.1667em\lower.7ex\hbox{E}\kern-.125emX}}

\begin{document}

\title{Automated Extraction of Acronym-Expansion Pairs from Scientific Papers}

\author{\IEEEauthorblockN{Izhar Ali}
\IEEEauthorblockA{\textit{Computer Science Department} \\
\textit{Rowan University}\\
Glassboro, NJ, USA \\
aliizh94@students.rowan.edu}
\and
\IEEEauthorblockN{Million Haileyesus}
\IEEEauthorblockA{\textit{Computer Science Department} \\
\textit{Rowan University}\\
Glassboro, NJ, USA \\
hailey74@students.rowan.edu}
\and
\IEEEauthorblockN{Serhiy Hnatyshyn}
\IEEEauthorblockA{\textit{Department of Bioanalytical Sciences} \\
\textit{Bristol-Myers Squibb}\\
Princeton, NJ, USA \\
serhiy.hnatyshyn@bms.com}
\and
\IEEEauthorblockN{Jan-Lucas Ott}
\IEEEauthorblockA{\textit{Department of Bioanalytical Sciences} \\
\textit{Bristol-Myers Squibb}\\
Princeton, NJ, USA \\
jan-lucas.ott@bms.com}
\and
\IEEEauthorblockN{Vasil Hnatyshin}
\IEEEauthorblockA{\textit{Computer Science Department} \\
\textit{Rowan University}\\
Glassboro, NJ, USA \\
hnatyshin@rowan.edu}
}

\maketitle
\begin{abstract}
This project addresses challenges posed by the widespread use of abbreviations and acronyms in digital texts. We propose a novel method that combines document preprocessing, customized regular expressions, and a large language model, specifically GPT-4, to identify abbreviations and map them to their corresponding expansions. The regular expressions alone are often insufficient to extract expansions, at which point our approach leverages GPT-4 to analyze the text surrounding the acronyms. By limiting the analysis to only a small portion of the surrounding text, we mitigate the risk of obtaining incorrect or multiple expansions for an acronym. There are several known challenges in processing text with acronyms, including polysemous acronyms (those with multiple meanings), non-local acronyms (those lacking explicit expansions nearby), and ambiguous acronyms (whose full forms do not correspond to the acronym letters). Our approach enhances the precision and efficiency of NLP techniques by addressing these issues with automated acronym identification and disambiguation. This study highlights the challenges of working with PDF files and the importance of document preprocessing. Furthermore, the results of this work show that neither regular expressions nor GPT-4 alone can perform well. Regular expressions are suitable for identifying acronyms but have limitations in finding their expansions within the paper due to a variety of formats used for expressing acronym-expansion pairs and the tendency of authors to omit expansions within the text. GPT-4, on the other hand, is an excellent tool for obtaining expansions but struggles with correctly identifying all relevant acronyms. Additionally, GPT-4 poses challenges due to its probabilistic nature, which may lead to slightly different results for the same input. Our algorithm employs preprocessing to eliminate irrelevant information from the text (i.e., authors' names, formulas, references, etc.), regular expressions for identifying acronyms, and a large language model to help find acronym expansions to provide the most accurate and consistent results. Overall, this work facilitates the creation of automated tools for extracting and expanding acronyms, thereby enhancing the readability and comprehension of scientific and technical documents.
\end{abstract}


\section{Introduction}
In today's digital age, where vast amounts of data are generated daily - measured in exabytes - the importance of natural language processing (NLP) and text mining becomes increasingly apparent \cite{b1, b2}. The widespread usage of abbreviations, specifically acronyms and initialisms, poses challenges for text comprehension and readability. A comprehensive analysis of 24 million article titles and 18 million abstracts spanning from the 1950s to 2019 revealed a 243\% and staggering 925\% increase of acronym use in titles and in abstracts, respectively \cite{b3}. Remarkably, over 94\% of all potential three-letter acronym combinations have been used at least once \cite{b3}. This work also uncovered that, out of the 1.1 million unique acronyms analyzed, 30\% appeared only once, 49\% were used between two and ten times, and merely 0.2\% of acronyms (slightly over 2,000) were cited more than 10,000 times \cite{b3}. This indicates that most defined acronyms are rarely reused in scientific documents.

Acronyms are a subset of abbreviations, which also include initialisms, truncations, contractions, etc. \cite{b13, b4}. Our work focuses solely on acronyms and initialisms. An initialism is a sequence of letters pronounced individually, for example, ``USA," ``FBI," ``CEO," and ``FAQ." An acronym, on the other hand, is typically formed by combining the first letters of multiple words and pronounced as a single word such as ``NASA," ``RAM," ``RADAR," and ``PIN." While some acronyms such as ``NASA" and ``USA" are universally recognizable, there are many domain-specific acronyms such as ``LPAR" (Logical PARtition) and ``RTEC" (RunTime Error Checking) that are less recognizable \cite{b6}. Additionally, nested acronyms such as ``JASA" (Joint Airborne SIGINT Architecture, where SIGINT stands for Signals Intelligence) and recursive acronyms like ``GNU" (GNU's Not Unix) or ``PIP" (PIP Install Packages) pose even greater comprehension challenges \cite{b6}. Acronyms can also exhibit polysemy, the ability of a single acronym to have multiple expansions within or across domains. For example, in medicine, ``ED" could denote ``eating disorder," ``elbow disarticulation," or ``emotional distress" \cite{b7}. On the other hand, ``TCP" has different meanings across various domains: in networking ``TCP" stands for ``Transmission Control Protocol," but in robotics, it signifies ``Tool Center Point."

Historically, acronym expansion mechanisms used manually crafted rules to identify acronyms and their full forms \cite{b8}. We are proposing a new approach by combining document preprocessing and text analysis, pattern recognition through regular expressions, and the advanced capabilities of large language models (LLMs), such as GPT-4 \cite{b41}, to provide a comprehensive tool for acronym identification and disambiguation. We aim for the identification and expansion of all acronyms in the provided digital texts, while mitigating known issues such as excessive acronym usage and Redundant Acronym Syndrome (RAS) \cite{b30}.

The rest of this paper is organized as follows. Section~\ref{sec:related_work} provides a brief overview of related work in the field of acronym identification and expansion. Section~\ref{sec:approach} outlines the proposed approach for identifying and expanding acronyms. Section~\ref{sec:design} focuses on the design and implementation of our algorithm while section~\ref{sec:challenges} discusses implementational challenges. In section~\ref{sec:results}, we present the experimental results and their analysis. We conclude our findings and outline directions for future work in section~\ref{sec:conclusion}. Throughout this paper, we will use ``short form" or ``acronym" to refer to the abbreviated form of an acronym, ``expansion" or ``full form" to denote the expanded form of an acronym, and an ``acronym-expansion pair" to denote an acronym together with its full form. Additionally, we used the ``4o mini" version of the GPT-4 model for this project, so we will use ``GPT-4" and ``GPT-4o mini" interchangeably.

\section{Related Work}\label{sec:related_work}
Pustejovsky et al. \cite{b9} used custom-designed regular expressions for identifying abbreviations in medical texts, highlighting the potential for rule-based systems in specific domains. Yeates et al. \cite{b6} introduced a novel compression method to validate acronyms based on surrounding text context. This method leverages the inherent patterns in language to infer possible expansions, offering an early example of context-based acronym resolution. Schwartz and Hearst \cite{b14} utilized heuristics to detect abbreviations presented within parentheses, a common convention in scientific literature. Their methodology has become a baseline for subsequent abbreviation detection algorithms due to its simplicity and effectiveness. Navigli and Velardi \cite{b17} expanded the scope of acronym detection through pattern matching and graph analysis. Their work illustrates the power of combining linguistic patterns with structural data analysis for acronym identification. Chang et al. \cite{b7} employed a combination of linear regression and dynamic programming for an online abbreviation dictionary. 

Nadeau et al. \cite{b15} were among the first to combine rule-based candidate generation with machine learning for validation, a hybrid approach that balances the strengths of rule-based systems with the adaptive learning capabilities of machine learning models. Chunguang et al. \cite{b11} utilized BERT-based models, Joopudi et al. \cite{b12} explored convolutional neural networks, and Nadeau et al. \cite{b15} applied supervised learning models, each contributing to the evolving landscape of acronym detection and validation. Jie et al. \cite{b18} and Cheng et al \cite{b26} proposed innovative approaches based on pronunciation and pattern recognition, respectively, while Ciosici et al. \cite{b19} and Haviv et al. \cite{b20} investigated the use of unsupervised learning and transformer language models.

These approaches underscore the potential of machine and deep learning in optimizing acronym-expansion extraction \cite{b11, b12, b15, b19, b20, b23, b28, b32, b33, b36}. However, the exploration of these advanced machine and deep learning techniques falls outside the scope of our study.

\section{Proposed Approach}\label{sec:approach}
Our approach to identifying and expanding acronyms from digital documents consists of three main components: (1) Document Preprocessing, (2) Regular Expression-Based Parser, and  (3) GPT-4 API Integration

\subsection{Document Preprocessing}
The first step of our methodology involves converting digital documents into plain text. We initially employed the Apache Tika toolkit \cite{b39} for PDF to text conversion, but we have transitioned to using PyPDF \cite{b43} for its enhanced compatibility with our processing workflow. Both tools yielded comparable outcomes in terms of the quality of text extraction and processing speed \cite{b44}. However, PyPDF offers a more streamlined integration with our workflow, eliminating the need for Java installation, which is a prerequisite for Apache Tika.

The plain text then undergoes standard NLP preprocessing steps such as tokenization, noise removal, and lemmatization \cite{b21, b23, b24, b25}. These steps are crucial for preparing the text for further analysis. NLP preprocessing steps are performed with an emphasis on maintaining the integrity and distinctiveness of acronyms. Specifically, we correct spelling errors and remove stopwords - common words with minimal informational value \cite{b24, b25} such as `a', `and', `the'. Our preprocessing meticulously preserves punctuation and maintains case sensitivity, critical for the accurate identification of acronyms, which often include capital letters and may contain hyphens. For spelling corrections, we only modify words that are recognized in the English dictionary and exhibit minor errors, such as an extra or missing character. This selective process ensures that acronyms, which may not align with standard dictionary entries and often feature unique combinations of letters, are not mistakenly altered.

\subsection{Regular Expression-Based Parser}
Our parser is designed to identify acronyms and their expansions within scientific literature, specifically targeting common patterns like ``acronym (expansion)" and ``expansion (acronym)" \cite{b9}. The parser efficiently detects acronyms that conform to these standard patterns and retrieves their expansions. When an acronym has multiple expansions as identified by these patterns, our parser preserves each of these expansions in the resulting analysis. This ensures a comprehensive representation of the acronym's various meanings.

The parser is unable to identify acronyms that deviate from these established patterns; particularly when an acronym and its expansion are in close proximity (i.e., within a sentence from one another) but both of them are missing contextual clues such as parentheses. For instance, the parser can identify the acronym BERT in \emph{``BERT stands for Bidirectional Encoder Representations from Transformers,"} but fails to capture the full expansion due to the absence of parentheses. We explored the idea of enhancing our regular expressions by incorporating keywords such as ``stands for," ``defined as," ``abbreviated as", etc., but it yielded a limited success. Although these enhanced regular expressions capture the expansion for BERT in \emph{``BERT stands for Bidirectional Encoder Representations from Transformers,"} they still fail with more complex instances such as \emph{``AIX, first released in the late 1980s, was IBM's advanced UNIX operating system."} Here, the pair \emph{``AIX: IBM's UNIX"} presents a challenge for regular expressions. However, by providing the acronym AIX along with the sentence as context to the GPT-4 model, it can effortlessly identify this pair. Despite our attempts to design specialized patterns for these situations, the inherent variability and absence of standard markers make reliable identification of acronym expansions challenging. The variability in how acronyms are introduced often surpasses the pattern-recognition capabilities of regular expressions, particularly for such non-standard cases. Additionally, attempting to define such regular expressions risks capturing incorrect or incomplete expansions, complicating the extraction process for LLMs.

\subsection{GPT-4 API Integration}
After we process the text through our parser, we generate a Python dictionary with acronyms as keys. The dictionary values are either the acronym expansions (i.e., if successfully extracted), or the context surrounding the acronym. We define context as the sentence containing the acronym and the preceding sentence. Through trial and error, we discovered that including the preceding sentence leads to better identification of the acronym's expansion as opposed to including the sentence that follows the acronym. By omitting the following sentence, we reduced the number of tokens fed into GPT-4.

Using GPT-4 for acronym extraction and expansion directly (i.e., by feeding a PDF file without any pre-processing) showed to be ineffective in our trials. In our tests of ChatGPT interface with the GPT-4o mini model, we noted that the model failed to identify all acronyms and provided incorrect expansions in some instances. We verified the accuracy of the output by examining the models' results by hand. These observations are detailed in Table \ref{tab2}.

Based on our observations, we adopted a two-tiered approach. We first ran our custom parser on the text, sentence by sentence, using regular expressions to identify acronyms. For each identified acronym, we then provided GPT-4 (via API) with either the direct expansion or the acronym's context for focused analysis. The acronym's context was limited to two sentences. This method allowed GPT-4 to process each acronym within a manageable context, enhancing overall performance and improving the accuracy of extracted expansions. We compared the outcomes of using solely the GPT-4 model, just our regex parser, and a combination of both. Presented in the results section, our findings clearly show that the integrated approach yielded the best performance. The high-level algorithm and the overall outline of our approach are shown in Figure \ref{fig1} below.

\begin{figure}[htbp]
\centerline{\includegraphics[width=\columnwidth]{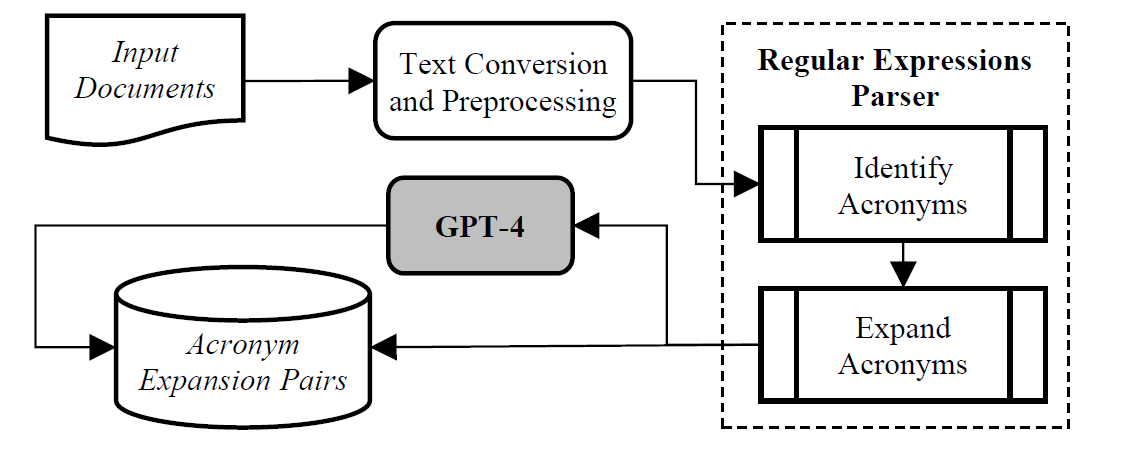}}
\caption{Approach for extracting acronym-expansion pairs.}
\label{fig1}
\end{figure}

\section{Design and Implementation}\label{sec:design}
This section provides a detailed overview of our acronym extraction algorithm, with an emphasis on the preprocessing steps that are critical for optimizing the document as input to GPT-4 for further analysis.

\subsection{Document Preprocessing}
We initiate the preprocessing phase by converting PDF documents into plain text using PyPDF \cite{b43}. Following the conversion, we undertake a series of steps \cite{b3} aimed at preparing the text for acronym-expansion extraction:

\begin{itemize}
    \item Remove headers and footers. Read each page via PyPDF, noting the first and last lines. If these lines repeat across pages, identify them as headers or footers and remove them from the text.
    
    \item Remove titles, abstracts, stopwords, math symbols, and equations.
    
    \item Remove the references section as it can include a mix of acronyms not directly relevant to the document's core content.
    
    \item Remove Roman numerals ranging from I to XXX. The upper limit of XXX is typically encountered based on conclusions from Meta research \cite{b3}.
    
    \item Remove chromosome formulas such as XX, XY, XO, ZO, XXYY, ZW, ZWW, XXX, XXXX, XXXXX, YYYYY, which often appear in biology and chemistry manuscripts.
    
    \item Remove gene sequences defined by six or more characters containing only A, T, C, G, and U, to avoid mistaking them for acronyms.
    
    \item Exclude acronyms longer than 10 characters. This is to avoid confusion with gene sequences or other non-acronym strings.
    
    \item Ignore strings preceded by numbers (e.g., 12-ACG) which typically represent chemical compounds or measurements rather than acronyms.
    
    \item Exclude headings and sub-headings that are often formatted in uppercase and could be mistakenly identified as acronyms.
    
    \item Replace ligatures. Ensure that typographic ligatures (where two or more characters are combined into a single glyph) are replaced so that character combinations are correctly interpreted.
    
    \item Remove excess spacing and correct hyphenation artifacts. Focus on removing lines, tabs, and fixing issues particularly in two-column layouts, to improve text readability and processing accuracy.
\end{itemize}

Each of these preprocessing steps is performed to ensure that the text is optimally prepared for the subsequent stages of acronym identification and extraction.

\subsection{Regular Expression-Based Parser}
The identification and extraction of acronym-expansion pairs is the central part of our algorithm.

For \textbf{\underline{acronym identification}}, we utilize a regular expression defined by the code snippet (\ref{code:regex}) to pinpoint acronyms within the text:

\begin{lstlisting}[language=Python, caption={Regular expression for acronym identification}, label={code:regex}]
pattern = r"\b[A-Z][A-Za-z-]*[A-Z]s?\b"
\end{lstlisting}

The regular expression in code snippet (\ref{code:regex}) is explained below:

\begin{itemize}
    \item \texttt{\textbackslash b} marks a word boundary and ensures that the pattern starts at the beginning of a word.
    
    \item \texttt{[A-Z]} matches the first uppercase letter, which is typically the starting character of an acronym.
    
    \item \texttt{[A-Za-z-]*} allows for a sequence of any combination of uppercase letters, lowercase letters, or hyphens.
    
    \item \texttt{[A-Z]} requires that the sequence ends with an uppercase letter, maintaining the structure typical of acronyms.
    
    \item \texttt{s?} optionally matches an `s' at the end, accommodating plural forms of acronyms.
\end{itemize}

This pattern is designed to capture sequences of capitalized letters, which may include lowercase letters or hyphens and optionally end with an `s'. It effectively identifies acronyms such as LPARs, LC-MS, and GNCS-INdAM, while filtering out instances that do not conform to the expected format of acronyms, such as ``eLisp (Emacs Lisp)" or ``2FA (two-factor authentication)" that start with a lowercase character and a digit, respectively. We discovered that this method could inadvertently recognize capitalized terms that are not acronyms, such as certain chemical names or country codes. To address this, we mark the position of each identified acronym within the text, allowing for the subsequent extraction of contextual information for more accurate processing.

For \textbf{\underline{acronym-expansion extraction}}, our parser employs two distinct regular expression patterns, named ``Forward Pattern" and ``Backward Pattern".

\textit{\underline{Forward Pattern}} identification targets scenarios where the acronym is directly followed by its expansion placed within parentheses: \texttt{<acronym (expansion)>}, as shown in the code snippet (\ref{code:forward_pattern}).

\begin{lstlisting}[language=Python, caption={Forward pattern for pair extraction}, label={code:forward_pattern}]
forward = r"\b" + re.escape(acronym) + 
          r"\b\s*\(((?:\b[a-zA-Z]\w*(?:-\w+)*\b\s*)+)\)"
\end{lstlisting}

\begin{itemize}
    \item \texttt{re.escape(acronym)} adds escape characters to all special characters in a given acronym, ensuring that the acronym is treated as literal text.
    
    \item \texttt{\textbackslash s*} allows optional whitespace after the acronym.
    
    \item \texttt{\textbackslash (} and \texttt{\textbackslash )} identify the opening and closing parentheses.
    
    \item \texttt{((?:\textbackslash b[a-zA-Z]\textbackslash w*(?:-\textbackslash w+)*\textbackslash b\textbackslash s*)+)} captures the acronym expansion.
    \begin{itemize}
        \item \texttt{\textbackslash b[a-zA-Z]\textbackslash w*} matches words that start with an alphabetic character: \texttt{\textbackslash b} ensures the word starts at a boundary; \texttt{[a-zA-Z]} matches the first letter; and \texttt{\textbackslash w*} matches the rest of the word, including letters, digits, or underscores.
        \item \texttt{(?:-\textbackslash w+)*} matches hyphenated words by using a non-capturing group: \texttt{(?: …)} groups a hyphen and following characters; \texttt{-\textbackslash w+} matches a hyphen followed by one or more word characters, allowing for the capture of terms like ``state-of-the-art" or ``Machine-learning-based."
        \item \texttt{\textbackslash b\textbackslash s*} allows for optional spaces between words.
        \item \texttt{+} ensures that the entire sequence (words and optional hyphens/spaces) can repeat, allowing for the capture of multi-word expansions.
    \end{itemize}
\end{itemize}

This pattern effectively extracts pairs like ``AIX (IBM’s UNIX)" and ``PHYP (IBM’s hypervisor for POWER systems)" given the text \emph{``... using AIX (IBM’s UNIX) and PHYP (IBM’s hypervisor for POWER systems) ...".} However, this pattern might misinterpret the content within parentheses. For instance, in the text \emph{``we applied LC-MS (which is usually used in practice) ...",} the text inside the parentheses is not the expansion of LC-MS. To mitigate this, we implement a stopword-based validation process \cite{b22} to filter out extracted expansions with a high proportion of stopwords. The validation process examines the proportion of stopwords in the expansion and if it exceeds a preset threshold, the expansion is discarded.

\textit{\underline{Backward Pattern}} identification is utilized when the expansion is followed by the acronym placed in parentheses: \texttt{<expansion (acronym)>}, as shown in the code snippet (\ref{code:backward_pattern}).

\begin{lstlisting}[language=Python, caption={Backward pattern for pair extraction}, label={code:backward_pattern}]
backward = r"\s((?:\b" + acronym[0] + r"[a-zA-Z\s-]*" + 
           acronym[-1] + r"[a-zA-Z\s-]*)+)\s*\(\b" + 
           re.escape(acronym) + r"\b\)"
\end{lstlisting}

The highlights of the key parts of the regular expression defined by the code snippet (\ref{code:backward_pattern}) are provided below (we omitted explanations of the parts that are also used in code snippets 1 and 2):

\begin{itemize}
    \item \texttt{\textbackslash s} looks for a whitespace character at the start. It ensures the expansion starts as a separate word and not in the middle of another word.
    
    \item \texttt{(?:\textbackslash b" + acronym[0] + r"[a-zA-Z\textbackslash s-]*)"} matches the first letter of the acronym at a word boundary, followed by any number of letters, spaces, or hyphens.
    
    \item \texttt{acronym[-1] + r"[a-zA-Z\textbackslash s-]*"} ensures the sequence ends with the last letter of the acronym.
\end{itemize}

The pattern defined by code snippet (\ref{code:backward_pattern}) effectively identifies correct expansions but may include extraneous words. To refine overly lengthy expansions, we reapply the backward pattern if the expansion's word count exceeds the acronym's character count, ensuring a closer match to the acronym's actual meaning.

For example, in the text \emph{``... a large language model (LLM) ..."} the backward regular expression pattern will identify ``large language model" as the expansion of LLM. In some cases, the initial results produced by the backward pattern are too lengthy, capturing correct expansions along with extra words. For example, in the text \emph{``... carbon samples secondary chemical shifts (SCS) ...",} our algorithm will initially identify \emph{``sample secondary chemical shifts"} as a full form of SCS. In such cases, we re-apply the backward regular expression to further refine the results to \emph{``secondary chemical shifts"} for SCS. The refinement is initiated only when the word count in the expansion exceeds the total number of characters in the acronym.

\subsection{GPT-4 API Integration}
Following the detailed processing by our regular expression-based parser, we proceed to the subsequent crucial phase of our methodology. We gather all acronyms, along with their expansions or contextual sentences, and feed them to the GPT-4 model for further refinement through a carefully engineered prompt (see code snippet \ref{code:prompt}). The prompt is designed to instruct GPT-4 to check the accuracy and conciseness of acronym expansions, and infer expansions based on the context when necessary. For acronyms that already have associated expansions, GPT-4 evaluates these expansions, making adjustments as needed to ensure they are precise and coherent. In cases where our parser does not successfully extract an expansion, GPT-4 analyzes the provided context to infer the most accurate expansion.

The model is also directed to exclude irrelevant information such as author names and proper nouns that are not acronyms. The interaction with GPT-4, conducted through its API, yields responses in JSON format, which allows for seamless integration into our existing processing pipeline. This step significantly enhances the overall robustness and reliability of our approach by ensuring consistent outputs that minimize manual post-processing efforts.

\begin{lstlisting}[caption={Python prompt for GPT-4 acronym refinement}, label={code:prompt}]
def get_prompt(self, text):
    prompt = f"""
    As an AI language model, you are tasked with refining a dictionary of acronyms and their explanations provided below:
    {text}

    Please follow these instructions carefully:

    1. Each entry in the dictionary consists of an `ACRONYM` and its corresponding `value` (full form or context).
    2. The `value` may contain the full form of the acronym or a context in which the acronym is used.
    3. If the `value` does not start with "(context)", check the accuracy and conciseness of the full form and make adjustments as necessary.
    4. If the `value` starts with "(context)", the full form of the acronym should be extracted based on the context provided.
    5. If the full form cannot be determined from the context, use your best judgment to provide the most accurate and concise full form.
    6. If you cannot determine the full form from the context, ignore the entry.
    7. Ignore author names, publication titles, locations, roman numerals, and other proper nouns that are not acronyms.

    Your output should be an updated dictionary in JSON format, adhering to the following structure:
    {{
        "ACRONYM": "Full Expansion of the Acronym",
        "ANOTHER_ACRONYM": "Full Expansion of Another Acronym",
        ...
    }}

    Ensure the final dictionary is accurate, concise, and formatted correctly for JSON compatibility. Exclude any additional text, comments, notes, or explanations outside of the updated dictionary entries.
    """
    return prompt
\end{lstlisting}

\section{Implementation Challenges}\label{sec:challenges}
During the design and implementation of our system, we encountered several challenges:

\textit{\underline{Handling GPT-4o mini's Input Limitation:}} The GPT-4o mini model has a 128k-token input limit, which posed challenges when working with large documents. To ensure comprehensive text analysis without losing important information \cite{b35}, we addressed this by breaking the text into smaller segments that fit within the limit for multiple API calls \cite{b29}. If the model is given inputs that exceed this limit, it can behave unpredictably \cite{b36}, as it starts extrapolating beyond its trained data distribution.

\textit{\underline{Computational Complexity:}} The inherent complexity of transformer models, particularly their attention mechanisms, presented another hurdle. The quadratic complexity of these mechanisms in relation to input length posed significant computational demands, especially for lengthy inputs \cite{b27, b28, b29}. Moreover, feeding the model with lengthy inputs can compromise the correctness and consistency of its output, leading to contradictions, digressions, or even exceeding the model's context capacity \cite{b37, b38}.

\textit{\underline{Optimizing Data Chunking:}} We evaluated the optimal size for data chunks to be processed by GPT-4o mini. Balancing the need to minimize API calls against the risk of exceeding input limitations, we found that presenting 15-20 acronyms along with their expansions or contextual sentences struck the best balance. This strategy facilitated efficient processing while adhering to the quality standards for expansion extraction.

\textit{\underline{API Rate Limits:}} GPT-4o mini's API rate limits required careful management to prevent excessive calls or token processing in a given timeframe. We optimized our algorithm for concurrent API interactions, ensuring adherence to the rate limits and equitable token distribution across calls. This approach maintained consistent quality in the extraction process, maximizing our utilization of GPT-4o mini's capabilities within operational constraints.

\textit{\underline{The Role of the Parser:}} Our regular expression-based parser was pivotal in overcoming these challenges. It ensured that inputs to GPT-4o mini were concise and relevant, thus:
\begin{itemize}
    \item Keeping within the 128k-token limit by selecting only essential acronym data for GPT-4o mini analysis.
    \item Supplying GPT-4o mini with targeted prompts, each containing acronyms and up to two contextual sentences, ensuring domain-specific processing and reducing computational demands.
    \item Performing all preprocessing tasks locally, the parser significantly expedited the data preparation phase, optimizing both time and computational resources.
\end{itemize}

\section{Results and Discussion}\label{sec:results}

Our evaluation of the acronym extraction algorithm was conducted on a diverse dataset of 200 scientific papers sourced from arXiv \cite{b42}, covering the following four domains: Biochemistry (BC), Systems Biology (SB), Computational Linguistics (CL), and Numerical Analysis (NA). We randomly selected and processed 50 papers from each domain to ensure a balanced and diverse corpus for analysis.

\subsection{Content Analysis}
To lay the groundwork for our study, we first conducted content analysis of the dataset. This involved using our regular expression-based parser to count acronyms and identify unique acronyms within the papers. We \emph{manually} reviewed 10 random papers from each domain, (i.e., 40 in total), verifying the parser's efficacy in matching the acronyms as outlined in the documents. We used NLTK's sentence and word tokenizer for accurate linguistic parsing. This initial analysis is captured in Table \ref{tab1}, presenting an estimate of the average number of acronyms per paper is each domain, alongside other textual characteristics such as average character, word, and sentence counts per paper across the different domains. This foundational analysis is essential for understanding the textual diversity we addressed in our algorithm's evaluation.

\begin{table}[htbp]
\caption{Content Analysis for 200 Papers from Four Paper Domains}
\begin{center}
\setlength{\tabcolsep}{9pt} 
\renewcommand{\arraystretch}{1.3} 
\begin{tabular}{|c|c|c|c|c|}
\hline
\diagbox[dir=SE,width=10em,height=4em]{\hspace{-1.2em}\textbf{Avg. per paper}}{\textbf{\hspace{1.2em}Domain}} & \textbf{\textit{BC}} & \textbf{\textit{SB}} & \textbf{\textit{CL}} & \textbf{\textit{NA}} \\
\hline
Acronyms & 61 & 39 & 27 & 21 \\
\hline
Character Count & 49,951 & 54,047 & 31,087 & 38,834 \\
\hline
Word Count & 8,722 & 9,263 & 5,799 & 8,008 \\
\hline
Sentence Count & 322 & 323 & 220 & 318 \\
\hline
\end{tabular}
\label{tab1}
\end{center}
\end{table}

\subsection{Summary of Study Results}
Our evaluation process was structured into five distinct cases, corresponding to the different approaches to acronym extraction: (1) regular expression-based parser without preprocessing (RegEx), (2) regular expression-based parser with preprocessing (RegEx+Pre), (3) GPT-4 without preprocessing (GPT), (4) GPT-4 with preprocessing (GPT+Pre), and (5) regular expression-based parser with preprocessing and GPT-4 (GPT+RegEx+Pre). Summary of the results is provided in Table \ref{tab2}.

\begin{table}[htbp]
\caption{Summary of the Results}
\begin{center}
\setlength{\tabcolsep}{2pt} 
\renewcommand{\arraystretch}{1.3} 
\begin{tabular}{|c|c|c|c|c|}
\hline
\textbf{} & \textbf{Approach} & \textbf{Total Acronyms} & \textbf{\% Expansions Found} & \textbf{Total * \%} \\
\hline
\multirow{5}{*}{\textbf{BC}} & GPT & 823 & 100\% & 823 \\
& GPT+Pre & 729 & 100\% & 729 \\
& RegEx & 3751 & 14.8\% & 555 \\
& RegEx+Pre & 3650 & 16.2\% & 591 \\
& \textbf{GPT+RegEx+Pre} & \textbf{3650} & \textbf{84.9\%} & \textbf{3100} \\
\hline
\multirow{5}{*}{\textbf{SB}} & GPT & 865 & 100\% & 865 \\
& GPT+Pre & 622 & 99.7\% & 620 \\
& RegEx & 2676 & 18.7\% & 500 \\
& RegEx+Pre & 2592 & 20.9\% & 542 \\
& \textbf{GPT+RegEx+Pre} & \textbf{2592} & \textbf{80.9\%} & \textbf{2098} \\
\hline
\multirow{5}{*}{\textbf{CL}} & GPT & 578 & 100\% & 578 \\
& GPT+Pre & 526 & 100\% & 526 \\
& RegEx & 1576 & 20.4\% & 321 \\
& RegEx+Pre & 1539 & 21.2\% & 326 \\
& \textbf{GPT+RegEx+Pre} & \textbf{1539} & \textbf{80.5\%} & \textbf{1239} \\
\hline
\multirow{5}{*}{\textbf{NA}} & GPT & 622 & 100\% & 622 \\
& GPT+Pre & 493 & 100\% & 493 \\
& RegEx & 1417 & 20.2\% & 286 \\
& RegEx+Pre & 1258 & 20.2\% & 254 \\
& \textbf{GPT+RegEx+Pre} & \textbf{1258} & \textbf{72.2\%} & \textbf{909} \\
\hline
\end{tabular}
\label{tab2}
\end{center}
\footnotesize{Note: GPT excels at coming up with expansions, as it is designed to always provide an answer. However, it often misses identifying acronyms. RegEx, on the other hand, excels at identifying acronyms but performs poorly when expanding them. The combined approach of GPT+RegEx+Preprocessing identifies and expands significantly more acronym-expansion pairs, showcasing the strength of the combined algorithm.}
\end{table}

As expected, all approaches that relied on GPT-4 to find acronym expansions yielded good results. In particular, approaches that relied solely on GPT-4 to identify acronym-expansion pairs were able to find nearly 100\% of the pairs. GPT-4 was unable to find acronym-expansion pairs only in three instances (over the whole dataset), when it misidentified manuscript text as an acronym. Specifically, GPT-4 only failed to find an acronym-expansion pair for the following: ``MICROCARD", ``R-package," and ``l-bin." 

The regular expression-based parser, without the support of GPT-4, was able to find acronym expansions only about 20\% of the time. This can be explained by the fact that regular expressions search for acronym expansions only within the manuscript itself. Furthermore, regular expressions search the manuscripts for acronym expansions that follow a specific format. Unfortunately, manuscripts in fields such as chemistry, biology, numerical linguistics, etc., often use acronyms without explicitly expanding them within the document, providing the acronym expansion outside the immediate proximity of the acronym, and seldom following a specific and consistent format for defining acronyms.

To validate the parser’s accuracy for identifying acronyms in the manuscript, we \emph{manually} examined randomly selected 10 papers from each domain (i.e., 40 papers total). It took a person between 20 and 30 minutes to identify all acronyms in a paper. Because of such huge time demand for manual acronym identification, we were unable to collect statistics regarding the parser’s accuracy in identifying acronyms. Furthermore, lack of expertise in the specific domains (i.e., biochemistry, systems biology, etc.) hindered our ability to determine if identified sequences of characters are indeed valid acronyms within the domain. We had to rely on web search to check the validity of an acronym which only increased the duration of processing. By \emph{manually} examining 40 randomly selected papers, we were able to confirm that the parser does identify all acronyms in the paper that follow the regular expression pattern defined in code snippet (\ref{code:regex}). 

It is also necessary to highlight the importance of the PDF file preprocessing to remove information that could be misinterpreted as acronyms (i.e., equations, author names, information in headers and footers, etc.). Parsing PDF files has been shown to be notoriously challenging. We tested a variety of different preprocessing heuristics and were able to reduce the number of misidentified acronyms by the regular expression-based parser by up to 11\%. Improving the efficacy of preprocessing mechanisms will lead to a decrease in the number of misidentified acronyms. This, in turn, will likely increase the percentage of correctly found acronym expansions when using a combined GPT+RegEx+Pre approach since GPT-4 will no longer need to search for expansions of misidentified acronyms.

Finally, the most interesting scenario is when we combined GPT-4 with the regular expression-based parser and preprocessing. Our first observation is that the parser identifies significantly more acronyms than GPT-4 by itself. While, GPT-4 was able to find acronyms that do not follow the regular expression pattern (e.g., those that start with a lower case character or a number), it routinely misses a large number of acronyms identified by the regular expressions, such as NTP: Nucleoside Triphosphate, Cryo-EM: Cryo-Electron Microscopy, IDT: Integrated DNA Technologies,  TEV: Tobacco Etch Virus, etc. Surprisingly, GPT-4 correctly identified and extracted certain acronyms in some papers but not in others. For example, RNA : Ribonucleic Acid acronym-expansion pair was correctly identified in paper 450745v2.pdf but was missed in paper 442555v1.pdf. Both papers were from the biochemistry (BC) domain. At present, we do not have a good explanation for this phenomenon because of proprietary nature of the large language  models. 

\section{Conclusion and Future Work}\label{sec:conclusion}
This project began before LLMs became widely available. We quickly recognized that our parser alone cannot find all acronym expansions, as many scientific papers frequently use acronyms without expanding them. Initially, we considered building a web crawler to search the Internet for acronym expansions. However, the introduction of LLMs, such as GPT-4, rendered this approach obsolete. Our study showed that while GPT-4 is great at finding acronym expansions, it still struggles to consistently identify all relevant acronyms within scientific manuscripts. The challenges stem from the probabilistic nature of LLMs, which generate outputs based on statistical relationships in vast datasets. This probabilistic approach can lead to inconsistencies, where some acronyms are correctly identified in certain contexts but missed in others. Moreover, LLMs suffer from the ``lost-in-the-middle" issue where they tend to focus on the beginning and end of large prompts, often losing critical information in the middle. Since most of our prompts max out the context window of the model's input, the model might focus on the acronyms at the beginning and end while missing those in the middle.

As we progressed with the project, new versions of GPT-4 were introduced, prompting us to re-run our study multiple times using the latest iterations, including GPT-4o mini, released in May 2024. The current results were collected using this version. Throughout our testing, we've observed consistent performance improvements with each new version of GPT-4. It is possible that in the future, GPT-4 could become powerful enough to render our preprocessing and regular expression-based parser unnecessary. However, as it stands, the combination of our preprocessing steps and parser, alongside GPT-4, yields the best results. The preprocessing and parser effectively identify acronyms, while GPT-4 excels at finding their corresponding expansions.

This study underscores the challenges involved in parsing PDF documents. We experimented with various heuristics to filter out text elements that could be mistaken for acronyms such as mathematical formulas, references, headers and footers, Roman numerals, gene sequences, and chromosome formulas. Despite these efforts, the latest version of our preprocessing unit still may encounter occasional issues to accurately distinguish and remove this extraneous text. Improving the preprocessing unit's accuracy and efficiency has therefore become one of our top priorities.

Another limitation of our current approach is the parser's inability to identify acronyms that begin with a lowercase letter or a number. We are actively working on developing a new set of regular expressions to address this gap. Furthermore, we plan to expand our research by comparing the performance of our approach using other large language models, such as Claude, Gemini, and Llama, in addition to GPT-4. Exploring these alternatives may reveal that other models are better suited for the task of identifying acronyms and determining their expansions, potentially outperforming GPT-4 in this context.

\section*{Acknowledgment}

Our team would like to thank Bristol Myers Squibb for their support and funding of the acronym extraction project.

\bibliographystyle{IEEEtran}
\bibliography{references}

\end{document}